\documentclass[sigconf]{acmart}

\AtBeginDocument{%
  \providecommand\BibTeX{{%
    \normalfont B\kern-0.5em{\scshape i\kern-0.25em b}\kern-0.8em\TeX}}}

    \copyrightyear{2024} 
\acmYear{2024} 
\setcopyright{acmlicensed}\acmConference[HRI '24]{Proceedings of the 2024 ACM/IEEE International Conference on Human-Robot Interaction}{March 11--14, 2024}{Boulder, CO, USA}
\acmBooktitle{Proceedings of the 2024 ACM/IEEE International Conference on Human-Robot Interaction (HRI '24), March 11--14, 2024, Boulder, CO, USA}
\acmPrice{15.00}
\acmDOI{10.1145/3610977.3634976}
\acmISBN{979-8-4007-0322-5/24/03}

\author{Bengisu Cagiltay}
\email{bengisu@cs.wisc.edu}
\affiliation{%
  \institution{Computer Sciences Department, University of Wisconsin-Madison}
  \city{Madison}
  \state{WI}
  \country{USA}
}

\author{Bilge Mutlu}
\email{bilge@cs.wisc.edu}
\affiliation{%
  \institution{Computer Sciences Department, University of Wisconsin-Madison}
  \city{Madison}
  \state{WI}
  \country{USA}
}

\begin{document}

\title[Toward Family-Robot Interactions]{Toward Family-Robot Interactions: A Family-Centered Framework in HRI}


\renewcommand{\shortauthors}{Bengisu Cagiltay and Bilge Mutlu}

\begin{abstract}
As robotic products become more integrated into daily life, there is a greater need to understand authentic and real-world human-robot interactions to inform product design. Across many domestic, educational, and public settings, robots interact with not only individuals and groups of users, but also families, including children, parents, relatives, and even pets. However, products developed to date and research in human-robot and child-robot interactions have focused on the interaction with their primary users, neglecting the complex and multifaceted interactions between family members and with the robot. There is a significant gap in knowledge, methods, and theories for how to design robots to support these interactions.
To inform the design of robots that can support and enhance family life, this paper provides (1) a narrative review exemplifying the research gap and opportunities for family-robot interactions and (2) an actionable family-centered framework for research and practices in human-robot and child-robot interaction.

\end{abstract}

\ccsdesc[500]{Human-centered computing~Interaction design theory, concepts and paradigms}
\ccsdesc[500]{Human-centered computing~HCI theory, concepts and models}
\keywords{family-robot interactions, family-centered design, child-robot interaction, theory and methods}

\begin{teaserfigure}
    \centering
  \includegraphics[width=0.9\textwidth]{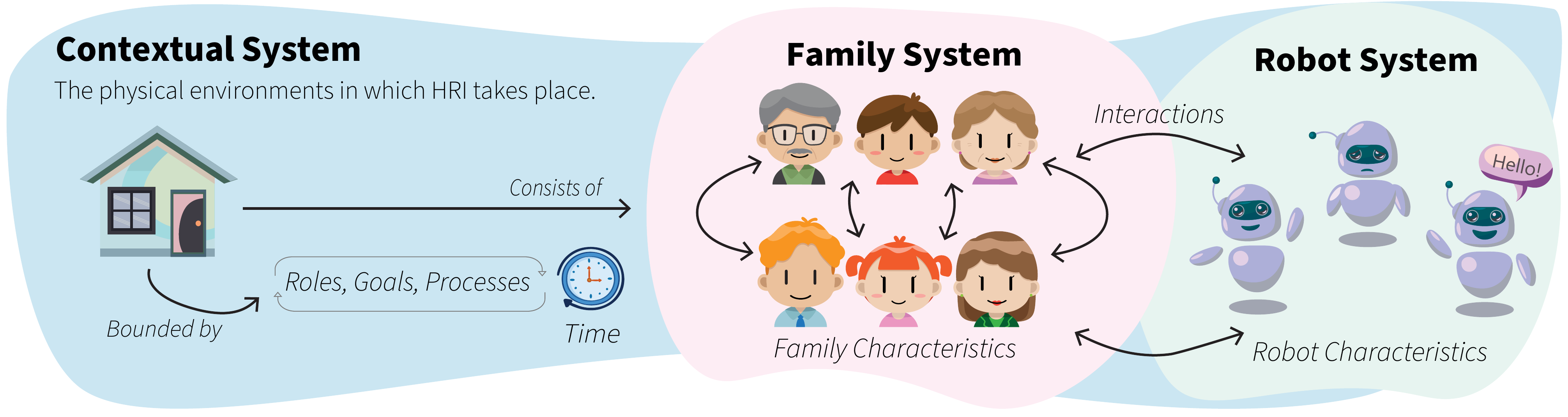}
  \caption{\textit{Toward Family-Robot Interactions:} The proposed family-centered framework bridges \textit{child-robot interaction practices} with \textit{family systems theories}. The framework includes three systems: Family system (e.g., family members and characteristics), robot system (e.g., its technical and interactive attributes), and the broader contextual system (e.g., the physical environment).}
  \Description{Collection of figures and text summarizing the three systems included in the framework. The contextual system is the physical environments in which HRI takes place, it is bounded by roles, goals, processes and time, and consists of the family system and the robot system. The family system includes a subtitle ``Family Characteristics'' and includes several images of family members and arrows connecting between them. The robot includes a subtitle ``Robot Characteristics'' and includes three robots, one which is waving, one looks sad, and one says 'hello'. The family system and robot system are connected with arrows, labeled `interactions'.}
  \label{fig:teaser}
\end{teaserfigure}

\maketitle

\section{Introduction}
Families and their surrounding care-ecosystems (e.g., relatives, friends, or their neighborhood and school) are important parts of children's development, embodying a rich space of shared experiences, values, and emotions. 
Research has shown that children and families' joint engagement with media and technology can support learning~\cite{holloway2019scaffolding}, entertainment~\cite{chambers2012wii}, building routines~\cite{sonne2016changing}, and achieving common daily tasks and goals~\cite{beneteau2020parenting}.
Introducing social robots into the broader context of families can potentially lead to similar positive outcomes. 
However, despite the field's movement toward more realistic, in-the-wild HRI, a \textit{family-centered approach} remains largely understudied in both theoretical and practical domains. 
So far, design paradigms for research in child-robot interaction often sideline the multi-layered relationships and differential expectations between children and family members (e.g., child-parent, child-sibling, grandparent-grandchild), typically focusing on children's individual interactions with robots (e.g., for therapy~\cite{scassellati2007social} or learning ~\cite{belpaeme2018social}). 
Family members are rarely active participants in interactions with robots, and typically serve as providers of consent or feedback. 
In recognition of this potential, a small but growing number of studies focus on robots designed with both the children and parents in mind (e.g., parent-child co-learning~\cite{ahtinen2023robocamp}, parent-child storytelling~\cite{chen2022designing}, or parent-child goal-oriented play~\cite{mchugh2021unusual}). 
Still, understanding human-robot interaction within a broader family context is challenging, due to the limited applications of current robot technologies to autonomously maintain multi-party and long-term interactions in domestic settings.
Overall, there is limited representation for any theoretical application that captures the broader domain of family needs, preferences, and interactions with robots. 

Given this gap, we aim to highlight the importance of exploring the theoretical foundations necessary to construct a robust family-centered design framework. In this work, we first review the small but emerging body of work in the HRI literature that considers children and families, we then bridge knowledge from family theories, and provide a theoretically motivated framework toward a \textit{family-centered approach in HRI}. Sampling from domains including healthcare, education, and emotional well-being, we synthesize the literature and demonstrate \textit{how interactions with robots can positively impact family life} (e.g., fostering family connections, bonding, co-learning, and dynamics) as well as \textit{how family members' characteristics may inform robot design} (e.g., robot perceptions, long-term adoption, and acceptance). We present a family-centered framework consisting of three components: the \textit{family system} (characteristics of and relationships between the family members involved); the \textit{robot system} (the robot's technical and interactive attributes); and the \textit{contextual system} (e.g., the physical environment in which interactions takes place). We illustrate the relationship between these systems and discuss how they are bounded by dimensions including \textit{time} (e.g., changes that happen over time) as well as the \textit{roles, goals, and processes} executed by each family member and robot.

By acknowledging the multi-layered relationships between family members, this framework seeks to enable the design of social robots that not only serve as functional aids for the individual but also integrate into the complex aspects of family life. 
Although we take a lens of domestic robots for families with children, we encourage researchers to extend this framework to other family structures and contexts that may be significant for family-robot interactions. 
Our contribution to this work is two-fold:
\begin{itemize}
\item A narrative review exemplifying the limitations and gaps in the current state of family-robot interactions,
\item An actionable theoretical framework for HRI researchers to consider when following a family-centered approach.
\end{itemize}

\section{Background}
The theoretical background of this work lies in the intersection of child-robot interaction, family theories, and ecological theories. 

\subsection{Child-Robot Interaction}
Research in child-robot interaction has demonstrated that robots engaging with children at home can positively influence their social and cognitive development. Robots can motivate children to undertake household chores, such as tidying their rooms~\cite{fink2014robot}, encourage reading habits~\cite{michaelis2019supporting}, and aid in emotion regulation~\cite{isbister2022design, slovak2018just}. Interactive agents can facilitate children's engagement in social play and foster positive bonds with robots through games and entertainment~\cite{belpaeme2012multimodal}. 
In such interactions, social robots may assume diverse roles, including as home assistants~\cite{de2005assessing, dautenhahn2005robot}, socially assistive agents for autism interventions~\cite{scassellati2018improving}, and educational interventions~\cite{belpaeme2018social}. Children and adolescents attribute roles like coaches, assistants, companions, and confidants to social robots~\cite{michaelis2018reading, cagiltay2020investigating, alves2022robots}. 
Social robots can even encourage family playtime and participation~\cite{kim2022can}. However, children and parents may have conflicting needs and expectations from in-home social robots~\cite{cagiltay2020investigating} as also seen in family interactions with smart home devices~\cite{beneteau2019communication, beneteau2020parenting, sun2021child}. Such conflicts may contribute to challenges that hinder sustaining long-term engagement with social robots and user acceptance~\cite{de2016long, de2015living, Cagiltay_engagement_2022}. Overall, child-robot interaction research has shown that robots in the home can positively impact children's social and cognitive development, motivate them in various tasks, support reading habits, help with emotional regulation, and serve as companions, assistants, and educational aids. However, there is little evidence on how robots can be used in the broader family context.

\subsection{Family Theories}\label{sec:theory}
Early applications of family theories draw from family clinical practices, e.g., Bowen's Family Systems Theory (FST)~\cite{bowen1966use}. FST argues that families form intricate social systems, wherein members continually influence each other's actions. Similar to biological systems~\cite{cox1997families}, family systems are argued to be more than the sum of its parts, capturing complex dynamics and capable of adaptive self-stabilization and self-organization in response to changes in the broader system~\cite{cox1997families}. Families consist of smaller subsystems which may capture interactions between parents, spouses, siblings, etc., as well as part of larger systems such as a neighborhood or community. 
From this perspective, a social robot residing in a household arguably becomes a member of the family system. It can form diverse connections and relationships with other members, exert influence over their behaviors, and, conversely, be influenced by their actions. The application of a family-systems approach in child-robot interaction research allows us to explore how robots can facilitate connections between parents and children, siblings, or across generations through shared social interactions~\cite{cox2010family}. A family-systems approach can also enable us to understand parental expectations, concerns, and acceptance of social robots within the home context~\cite{lin2020parental}. By integrating family-systems theories into the study of child-robot interactions (e.g., ~\cite{cagiltay2023family}), we can better inform the design of social robots to support the well-being of families.

\subsection{Ecological Theories}

Ecological theories provide holistic views for understanding how people interact with technology. The ``Product Service Ecology'' model proposed by ~\citet{forlizzi2013product} adapts the ``Social Ecological'' model, providing designers with a holistic perspective for understanding the contextual factors influencing product design and usage. ~\citet{forlizzi2006service} applies an ecological model to investigate how autonomous service robots, such as Roomba vacuum cleaners, integrate into family homes through ethnographic observations. Similarly, ~\citet{sung2010domestic} presents the ``Domestic Robot Ecology'' model to foster a comprehensive understanding of complex in-home settings and the acceptance of autonomous service robots for long-term use. Moreover, an ``Ecology of Aging'' approach explores how robots can be designed to assist the elderly at home~\cite{forlizzi2004assistive}. These applications mainly focus on domestic service robots, such as robotic cleaners, as opposed to social companion robots. 
%
Bronfenbrenner's bioecological theory~\cite{bronfenbrenner1979ecology, bronfenbrenner1998ecology, bronfenbrenner2007bioecological} describes four layers that shape an individual's development: individual's immediate relationships (microsystem); interactions between or within levels (mesosystem); relationships with indirect environments (exosystem); and societal or belief-level influences (macrosystem). These systems are influenced by processes (interactions with objects or people), personal attributes (interests, appearances), contexts (home, school, community), and time (historical changes and duration of processes). Family Ecological Model (FEM) \cite{andrews1981ecological} is an adaptation of this bioecological theory, that similarly argues that family interactions, such as parenting, are shaped by the context in which the families are situated~\cite{davison2013reframing}. Recognizing the significance of ecological theories, we aim to extend the application of these models to the design and development of a framework for family-robot interactions.

\section{Social Robots for Families: Review}
We conducted a literature review to investigate the current state of family-centered approaches within human-robot interaction. 
We specifically prioritized applications of \textit{social robots}, within \textit{real-world} settings, for \textit{families with children}. We focus on these three areas because they pose a particularly challenging context for robot design where actionable frameworks can be useful. Unlike service robots, social robots are capable of fostering emotional connections, rich interactions, and perceived as a member within a family setting, making them more relevant for exploring the impact of technology on family dynamics and relationships. 
Acknowledging that ``family-robot interactions'' is not yet a widely established term within HRI, we grounded our review within child-robot interactions as our primary keyword, hence focusing on ``social robots for families with children.'' From this, we expanded our search to explore the participation of broader family structures and real-world contexts in this evolving field.

\subsection{Research Questions}
We asked the following research questions:

\textbf{RQ1. Which family members are typically involved in CRI research?} 
\textit{(e.g., Siblings, mothers, fathers, grandparents, relatives)}

\textbf{RQ2. What are the roles of families in CRI research?}\\ 
\textit{(e.g., Active participant, observer, informant, translator, non-participant)}

\textbf{RQ3. What context are social robots used for children and families?}
\textit{(e.g., Education, healthcare, therapy, in-home, museum)}

\subsection{Search Method and Review Criteria}
We conducted a semi-structured narrative literature review to identify relevant studies and scholarly works in the intersection of \textit{family-centered human-robot interaction}. We searched from the ACM Digital Library and Scopus databases. Since our review is contextualized around ``families with children'' our initial querying approach focused on keywords surrounding social robots, families, and children or teens: TITLE OR ABSTRACT OR KEYWORDS (social AND robot*) AND (family OR families) AND (teen* OR child*). However, the search results yielded limited entries (ACM DL: 23, Scopus: 97). We then broadened the term ``family'' in our search to: TITLE OR ABSTRACT OR KEYWORDS (social AND robot*) AND (family OR families OR parent* OR mother* OR father* OR sibling*) AND (teen* OR child*). In total, this search returned 315 papers (ACM DL: 54, Scopus: 261). After excluding duplicates, non-English papers, and non-papers (e.g., proceeding reports) there were 248 remaining potential sources (See Figure \ref{fig:PRISMA}). This final query indexed all articles until April 16, 2023. 
Two authors participated in the review process and screened the potential 248 papers in three-steps:

\begin{figure}
    \centering
    \includegraphics[width=\columnwidth]{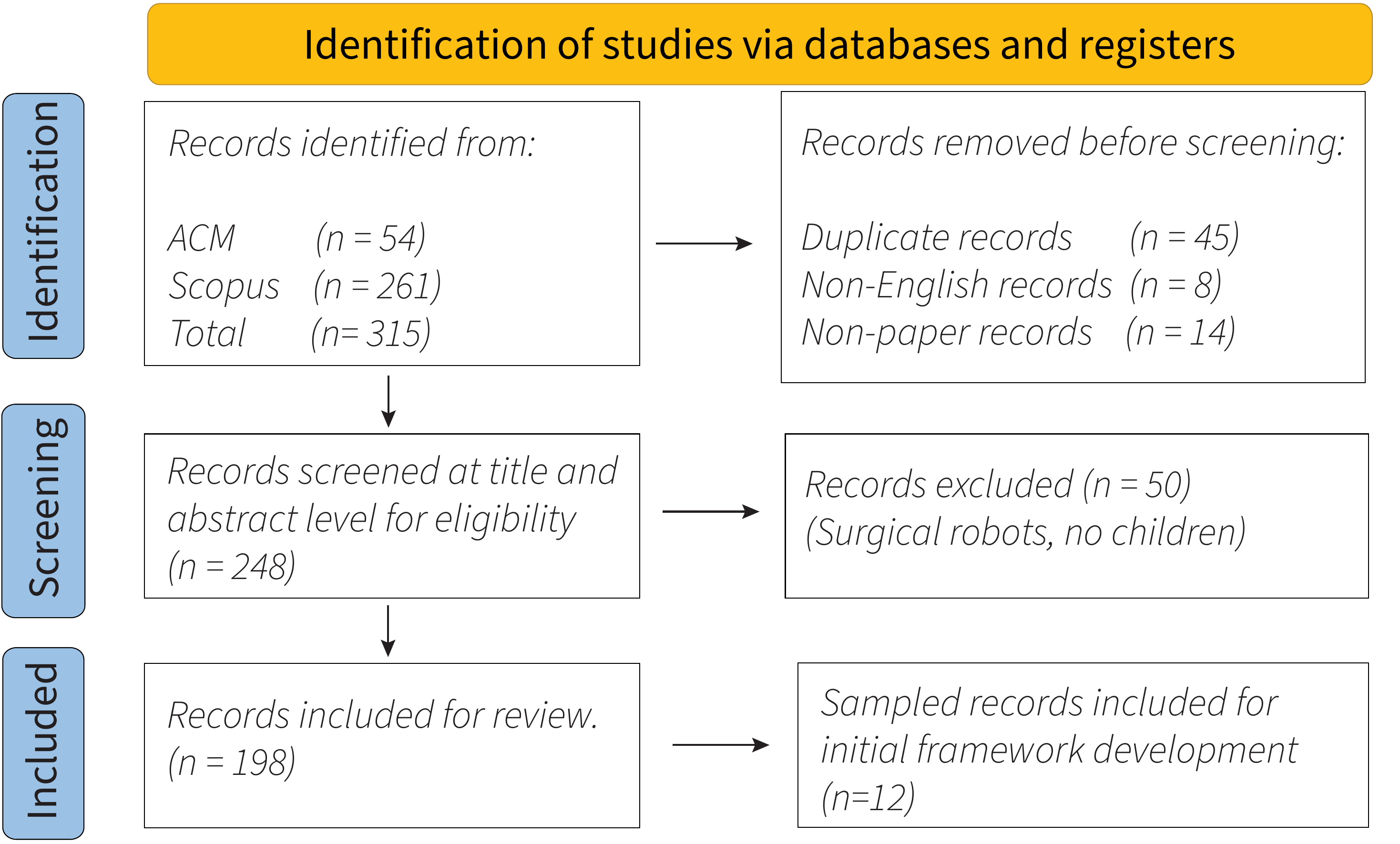}
    \caption{Search method and criteria for the review process.}
    \label{fig:PRISMA}
    \Description{ A table describing the search method and criteria for the review process. The table has three rows ``Identification'' ``screening'' and ``included'', and two rows describing the number of records in each column.}
\end{figure}
\subsubsection{Initial Screening} The first author conducted the initial paper screening. We included papers focusing on children's robot use with an emphasis on family involvement. Papers were excluded if they were not related to the HRI field and research questions (e.g., surgical robots in pediatrics). 50 papers were excluded.

\subsubsection{Secondary Screening} For the remaining 198 papers, the authors created a codebook and identified six categories for secondary screening: the domain of the study (e.g., healthcare, learning, telepresence); type of study (e.g., review, user study, position paper); contextual location of the study (e.g., lab, in-home, school, hospital); the role of the robot in the study (e.g., social companion, assistant, mediator); the type of robot used (e.g., Nao, Pepper, Cozmo etc.); which family members were included (e.g., parents, siblings); and the role of family members in the study (e.g., informer, observer, participant). The first author coded the 198 entries based on the codebook and revised the coded entries with the second author. Through iterative discussions, the entries were organized into thematic categories, allowing for the identification of common trends, patterns, and gaps in the literature. 

\subsubsection{Sampled Papers for Framework Development} After the secondary screening of 198 papers, the first author conducted a relevance scoring to determine the extent to which each paper aligned with the focus on family-robot interaction. Papers were scored high, mid, or low within the following categories: (1) Family: Papers that directly involved family members OR those that explored family-oriented contexts; (2) Robot: Papers that addressed the robot's ability to engage in social interactions within the family OR the perceived influence of the robot on family dynamics; and (3) Setting: Papers capturing naturalistic real-world settings OR long-term interactions. Papers that showed evidence in these categories were labeled high due to their relevance to understanding interactions in authentic family environments. 
At this step, 65 out of 198 papers were ranked high or mid in one or more of these three categories. Notably, in this work, we sampled a smaller subset of 12 papers that include studies where children and family members were involved as participants in interactions with a social robot in real world settings. We further synthesize these papers that showcase actionable and novel contributions which served as a feasible first step towards structuring our proposed framework\footnote{ \href{https://osf.io/gba6v/?view_only=09e726a02e874a6592c6e47d381c82d6}{The full list of selected papers can be found in this embedded repository link.}}.

\begin{table*}
    \centering
    \caption{Categories of the sampled papers used to construct an initial family-centered framework}
    \begin{tabular}{p{0.48\linewidth}p{0.48\linewidth}}
    \hline
        \textbf{Robot Interactions That Impact Family Life} & \textbf{Family Characteristics That Inform Robot Design} \\ \hline
        Family Connections~\cite{chen2022designing, abe2018chicaro}; At Home Special Education and Family Bonding~\cite{armstrong2021longing}; Family Co-Learning~\cite{ahtinen2023robocamp}; Family Dynamics~\cite{pelikan2019stubborn}  
        & Robot Perceptions~\cite{mchugh2021unusual}; Robot Adoption in Long-Term \cite{henry2020qualitative, amirova2023effects, levinson2022living}; Robot Acceptance \cite{de2021exploring, lin2021parental, lee2022can} \\ \hline
    \end{tabular}
    \label{tab:summary}
    \Description{A two column table listing the synthesis of sampled papers. Left column lists papers that were labeled ``robot interactions that impact family life'' and the right column lists papers that were labeled ``family characteristics that inform robot design''}
\end{table*}

\subsection{Descriptive Insights From Screened Papers}
To answer our three research questions, we provide a descriptive overview of the 198 papers within our search scope. 

Over half of the screened papers (54\%) included robot use in \textit{healthcare} settings for children. Applications included interventions for Autism Spectrum Disorder (ASD), chronic illnesses, sensory impairments, ER visits, and hospitalization. Notably, 68 of 105 papers in this category focused on the use of robots to support children with disabilities and special needs. These studies typically took place in hospitals and labs. 
%
Approximately 16\% of the papers explored robot use in \textit{social interactions and companionship}, including topics such as parenting support, collaborative play, connection-building, diary keeping, storytelling, and sleep hygiene. These studies occur in settings such as homes, labs, museums, and shopping malls.
%
Nearly 13\% of the papers focused on robots use in \textit{educational roles}, such as reading, language learning, math, tutoring, and speech therapy. These studies took place in settings like classrooms, labs, or science and technology clubs.
%
Nine percent of the papers examined topics within \textit{psychology}, including developmental robotics, joint attention, synchrony, attribution of intelligence, and culture. These studies took place in controlled laboratory studies.
%
Five percent of the papers addressed robot use for \textit{emotional wellbeing} within activities for pain and anxiety management, music therapy, loneliness reduction, stress buffering, and painting. These interventions occur in playrooms, homes, and labs.

Among the 198 papers, the most widely used methods are \textit{user studies} (44\%), which included field studies and long-term studies. This was followed by clinical or lab interventions and trials (28\%), new robot designs or prototypes (15\%), or review and survey papers (13\%). 
Notable family-centered robot prototypes include Arpi~\cite{sarmiento2023arpi}, Emobie~\cite{arnold2016emobie}, Buddy~\cite{kim2017buddy}, Edu~\cite{talami2021edu}, and Pabi~\cite{dickstein2011affordable}. For example, Edu~\cite{talami2021edu}, is designed as a companion for children in protective isolation units to connect and play with family.

\paragraph{Roles of Families in Child-Robot Interaction Research}
The screened papers typically mentioned insights only from children, or reported findings from children accompanied by: a parent; a sibling; a parent and other stakeholders such as classmates, educators, teachers, clinicians, nurses, or therapists. Rarely, papers included infant-mother pairs, adolescents, or child-grandparent pairs.
We identified \textit{roles of family members} categorized as: informer; feedback provider; interpreter; consent provider; survey filler; diary keeper; reporter; passive observer; active participant.
We captured that, family members, typically parents, functioned as ``survey fillers'' on behalf of their children and selves. In longitudinal studies, they have acted as ``diary keepers'' or ``reporters,'' responsible of tracking interactions and metrics for behavioral insights. Most often, family members were either ``consent providers'' or ``passive observers'' quietly observing interactions between their child and robot without actively influencing them. Family members were rarely ``active participants'' that partake in the interaction with the robot and child. 

\subsection{Identifying the Research Gap} \label{sec:gap}
We identified several gaps that limit a family-centered approach in human-robot interaction research. 
First, the \textit{context predominantly includes robots for healthcare applications and social interventions for children with neuro-developmental disorders.} Within these child-robot interaction studies, there are limited investigation on family dynamics and social robot interactions in different contexts.
Second, \textit{family members in child-robot interaction studies mostly involve parents, and specifically mothers.} Siblings, grandparents, or other diverse forms of family structures are underrepresented.
Third, there is limited research examining design needs for \textit{how robots can support long-term interactions in continuously evolving family roles and dynamics.} The evolution of roles over time are not captured in these studies. In fact, family members often take an indirect role in research and have passive involvement with the robot.

\section{Synthesis of Sampled Papers} \label{sec:synthesis}
We provide an in-depth review of the 12 selected papers synthesized in two categories: \textit{robot interactions that support family life}; and \textit{family characteristics that inform robot design} (See Table \ref{tab:summary}). 

\subsection{Robot Interactions That Impact Family Life}
Robot interactions in contexts such as education, companionship, storytelling, and social assistance \textit{can support} family connections, family dynamics, child development, and learning. 

\subsubsection{Robots for Eliciting Family Connections}
\citet{chen2022designing} studied the design of a robot-assisted storytelling interaction with 12 families of 3-7-year-old children in the context of parent-child-robot interaction. Robots were deployed in family homes for 3-6 weeks and remotely teleoperated for six 25-minute sessions of parent-child-robot story reading. The findings show that social robots can improve parent-child interactions. Particularly, the robot's participation in parent-child-robot interactions fostered togetherness, ``three of us'' feeling, and a sense of belonging during long-term storytelling. This suggests that the robot's storytelling facilitated parent-child bonding and created a unique sense of shared experiences and unity in this triadic setting. This study shows the potential of robots that can foster family bonding and lasting memories through collaborative activities.

\citet{abe2018chicaro} investigated a remote-operated mobile telepresence robot for supporting interactions between toddlers and their non-co-located families. In a playroom field trial, researchers tested the system's acceptance and usefulness. The study included 36 participants, including 19 toddlers (aged 0 to 3) and adult family members (parents or grandparents). Eight toddlers refused to play with the robot in the playroom, while eleven toddlers interacted with the robot. The remote teleoperated system increased opportunities for grandparent-grandchild interactions and promoted the sharing of daily life. Motivated by these interactions, grandparents paid special attention to their appearance during video interactions. This phenomenon highlights the potential of utilizing robotic telepresence to strengthen toddler-grandparent relationships and foster meaningful connections.

\subsubsection{Robots for Special Education and Family Bonding}
In a three-month autoethnographic study~\cite{armstrong2021longing}, a parent of a 10-year-old child with autism describes their long-term interactions and experiences with an in-home socially assistive robot. The robot's engagement in simple social interactions, such as greetings, fostered direct and indirect bonding between the parent and child and supported improvement in the child's social skills. Notably, the child exhibited knowledge transfer from the robot after five weeks of sustained engagement. To support this period, the parent and a speech pathologist created a customized curriculum to maximize the robot's educational potential. During the study, the child showed the robot affection, noted as a personal milestone for the parent. After two months of interaction, the child could express ``love'' by reciprocating the phrase \textit{``I love you''} to their mother \textit{through the robot}. These interactions show how therapeutic robot interventions can facilitate parent-child communication and emotional expressions.

\subsubsection{Robots for Supporting Family Co-Learning}
In a one-month in-home study with eight families (sixteen parents and sixteen children aged 6-15 years old) in Finland, \citet{ahtinen2023robocamp} explored how family members collaboratively learn about social robots at their homes. Family members interacted with a robot collaboratively by completing many open-ended, hands-on tasks to learn about social robots, their potential roles and tasks, programming, and design. Study methods included online family interviews and diaries, where parents and children filled the diaries together. Notably, the authors emphasize parents' access to data throughout the study. For example, researcher communications with children occurred through the supervision of parents, and families were provided transparency in privacy-preserving options for the robot. Families perceived the co-learning as a positive experience, where the findings suggest that the co-learning experience created the feeling of collaboration and ``togetherness'' between family members.

\subsubsection{Robot Roles in Family Dynamics}
A one-week in-home study by \citet{pelikan2019stubborn} examined how families react to non-lexical sounds from a Cozmo robot and explored how families with at least one 8-14-year-old child interacted with the robot. The preliminary findings show that families simplified their sentences to mimic the robot's sounds. This adaptation suggests that the robot's communication style influenced family members' language. Additionally, some families portrayed the robot differently, with parents calling it a ``stubborn-child'' and children calling it a ``buddy.'' These dual characterizations demonstrate potential in the robot's multifaceted role within the family dynamic, ranging from a challenging yet endearing agent to a friendly figure.

\subsection{Family Characteristics That Inform Robot Design}
Family members' characteristics, beliefs, values, or conflicts may impact their perceptions, interactions, and attitudes toward robots.

\subsubsection{Family Characteristics on Robot Perceptions}
\citet{mchugh2021unusual} examined how family characteristics affect children's perceptions of robots through a goal-oriented play study in a museum. The study findings identified a connection between parents' characteristics and children's perceptions toward robots. Specifically, the results showed that children whose parents had STEM backgrounds were more likely to attribute higher animacy to the robot. The study also found a link between parent-child communication and children's perceptions. Specifically, the amount that parents communicated about the robot directly influenced a child's perception of robot animacy. For the children of parents with STEM backgrounds, children gave commands to the robot assuming the robot had advanced perceptual and cognitive abilities. These findings suggest that family characteristics, communication dynamics, and children's perceptions are interconnected factors that affect human-robot interaction.

\subsubsection{Family Characteristics on Long-Term Robot Adoption}
In a study of hospitalized children using a telepresence robot for two weeks or more, \citet{henry2020qualitative} explored the robot's role in helping children maintain connections with their parents and siblings and retain their sense of belonging within the family structure. This exploratory study interviewed 17 onco-hematology patients (ages 7-25) and their parents about their home telepresence robot experiences. The robot enabled family members to remotely communicate with the hospitalized child, supporting patients and families. Patients primarily valued the robot's ability to bridge connections with siblings, \textit{``assume their role at home''}, and \textit{``continue to have a life that is as normal as possible.''} However, this robot connection became less effective for some families as their daily lives became more complicated. When the hospitalized child was sick or tired, they sometimes avoided robot contact with the family. This, in turn, led to feelings of sadness among the siblings, who were unable to connect during such moments. The changes in these dynamics demonstrate the complex relationship between technology, family relationships, and family members' emotional well-being, emphasizing the need for nuanced approaches to family-centered human-robot interactions in healthcare.

A 21-day multi-session robot-assisted autism therapy (RAAT) study by \citet{amirova2023effects} included 16 children aged 5-12 and their parents in a Kazakhstan hospital therapy center. The study examined whether parental involvement during autism interventions affected the efficacy of RAAT interventions. Findings suggest that, parental presence may encourage exploratory and playful behaviors for non-verbal children with severe autism, but may decrease verbal children's compliance and increase aggression. Proposed guidelines for RAAT suggest that researchers should consult with parents before therapy, define their roles, and allow their presence in initial sessions, especially for severe cases of autism. Parent-involved data collection measures may include questionnaires, parent-child relationship assessments, skill benchmarks, and pre- and post-tests to ensure effective therapy and long-term progress monitoring. This study shows how parental presence, children's characteristics, and child-parent relationships affect RAAT outcomes and can inform robot design for long-term parent-involved therapy sessions.

In a two-week study by
\citet{levinson2022living}, seven families with children aged 6-13 shared their experiences and concerns about living with a social robot. Children engaged in entertaining activities with the robot, such as storytelling and word-guessing games. The activities encouraged collaborative play and turn-taking between siblings, however children faced challenges in sustaining long-term engagement, which also created a burden on parents. The results suggest that contextual factors such as changes in family routines during summer vacation, may hinder long-term engagement.

\subsubsection{Family Characteristics on Robot Acceptance}
A study by \citet{de2021exploring} examined children's views on robots and their acceptance or rejection of them. 87 children (7–11-year-olds) interacted with a social robot during class and completed a questionnaire about their experiences. The findings show that three types of beliefs influenced children's decision-making. Utilitarian beliefs, such as the robot's ability to help with homework, chores, and teach jokes, contributed to fostering acceptance. Hedonic beliefs --the desire for companionship, emotional support, and play when no one else was available-- were identified as strong motivators for acceptance. Social beliefs, such as children who view robots as ``cool'' and ``unique'' were more likely to accept them in their homes. In contrast, normative beliefs about family members' disapproval or fear of siblings were identified as reasons for rejection. This study shows how family and surroundings influence children's normative beliefs about robots in the home. This nuanced understanding of children's beliefs reflects the multifaceted nature of their decision-making processes for accepting robots in home. 

In a qualitative design fiction study, \citet{lin2021parental} explored predictors for parents' acceptance of storytelling robots in their families, including 14 mothers and 4 fathers with children aged 2-5. The context of use affected the perceived appropriateness of introducing such a robot into their family. Parents' acceptance also was reliant on the robot's perceived agency and intelligence. Notably, parents were ambivalent about the robot's emotional expression, which may hinder adoption. Although this work did not include insights directly from children, it demonstrates the factors that parents consider when deciding whether to incorporate storytelling robots into their family interactions.

~\citet{lee2022can} surveyed working parents regarding their expectations and preferences for social robots' in childcare functions, such as socialization, education, entertainment, and consultation. The survey included dual-income mothers ($N=351$) and fathers ($N=273$) with single or multiple children (early childhood, aged 3-7, $N=204$; middle childhood, aged 8-12, $N=210$). Parents favored robots for childcare functions like socialization, entertainment, and consultation, but not education. Additionally, working parents' \textit{parenting styles} (e.g., family-oriented, work-oriented, noninterventional, dominant) influenced their expectations regarding social robots. Children's characteristics also influenced parents' preferences for a social robot, such that families with younger children were more likely to prefer social robots with counseling functions. However, parents of older children were more positive towards an entertainment robot. These findings illustrate how child and parent characteristics influence acceptance of robots at the home.

\section{Toward a Family-Centered Framework in HRI}
We employed a \textit{deductive approach} to guide the framework development grounded by (1) the ecological theories described in Section \ref{sec:theory}, (2) the gaps noted in Section \ref{sec:gap}, and (3) the synthesis of the sampled 12 papers in Section \ref{sec:synthesis}. 


First, by incorporating family-systems theories, specifically Bowen's Family Systems Theory (FST), into our theoretical framework, we aim to capture the dynamics of family interactions with robots. FST's perspective on families as complex social systems provides insights into the potential role of social robots as integral members of these systems. This lens enables the exploration of diverse interactions and relationships formed between social robots and family members. \textit{Through FST, we identified three systems for the proposed framework: Family System, Robot System, and Contextual System.}

Second, we translated practical insights grounded in the theoretical background, research gaps, and paper synthesis, to identify the multifaceted roles of social robots within family systems. The synthesis served as a guide for developing the proposed framework that represents the complexities of family-robot interactions. From robots fostering togetherness through storytelling to robots strengthening family bonds via telepresence, these studies offered tangible examples of social robot characteristics that might contribute positively to family dynamics. The system components aim to inform the design and deployment of social robots, ensuring their meaningful integration into diverse family structures and sustained interactions over time. We describe these systems below.

\subsection{Family-Robot Interaction Systems}
\begin{figure}
    \centering
    \includegraphics[width=\columnwidth]{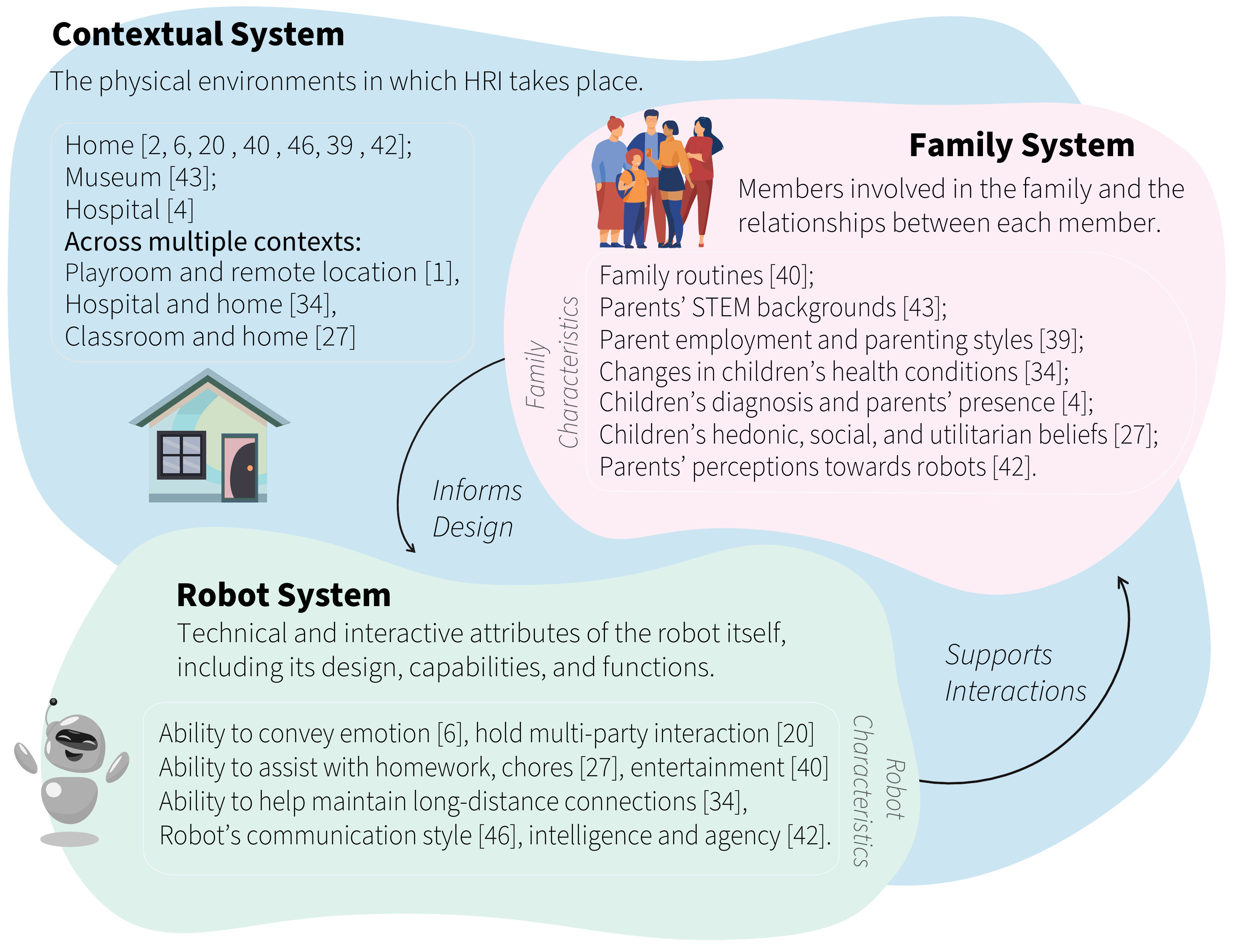}
    \caption{Systems for Family-Robot Interaction}
    \Description{This figure illustrates that Family-Robot Interactions consist of three systems: the contextual system, family system, and robot system. The papers that fall into each system and their categories are listed. There is an arrow from the family system to the robot system saying ``informs design'' and an arrow from the robot system to family system saying ``supports interactions''}    
    \label{fig:framework}
\end{figure}

At a high level, family-robot interactions consist of three systems: the family system, robot system, and contextual system (See Fig~\ref{fig:framework}).

\subsubsection{Family System}
The family system includes all members involved in the family and the relationships between each member. It considers the unique characteristics, beliefs, values, and conflicts within the family that influence their perceptions and interactions with robots. These members may be co-located or located in different contextual systems. The family system is dynamic and may evolve over time, impacting the long-term nature of robot interactions. For example, new members may join or leave the family system for various reasons (\textit{short-term changes:} work, vacation, school; \textit{long-term changes: }birth, death, marriage, divorce). 
From the literature review, we identified that several family members' characteristics play a significant role in shaping how family members perceive and engage with robots, these include: family routines~\cite{levinson2022living}; parents' STEM backgrounds~\cite{mchugh2021unusual}; dual parent employment and parenting styles~\cite{lee2022can}; family resilience towards changes in children's health conditions~\cite{henry2020qualitative}; the spectrum of children's diagnosis and parents physical presence~\cite{amirova2023effects}; children's hedonic, social, and utilitarian beliefs~\cite{de2021exploring}; and parents' perceptions towards the agency, intelligence, and expressivity of robots~\cite{lin2021parental}.

\subsubsection{Robot System}
The robot system considers the technical and interactive attributes of the robot itself, including its design, capabilities, and functions. We identified varying use cases of robots that have distinct impacts on family life, such as those designed for storytelling~\cite{chen2022designing, lin2021parental}, play~\cite{abe2018chicaro, mchugh2021unusual, levinson2022living}, learning~\cite{ahtinen2023robocamp, de2021exploring}, therapy~\cite{armstrong2021longing, amirova2023effects}, communication~\cite{pelikan2019stubborn, henry2020qualitative}, and childcare~\cite{lee2022can}.
Several robot characteristics may support interactions between family members, for example, robot's ability to convey emotion~\cite{armstrong2021longing}, hold multi-party interactions~\cite{chen2022designing}, assist with homework, chores~\cite{de2021exploring}, entertainment~\cite{ levinson2022living}, help maintain long-distance connections~\cite{henry2020qualitative}, as well as its communication style~\cite{pelikan2019stubborn}, perceived intelligence and agency~\cite{lin2021parental}.

\subsubsection{Contextual System}
The contextual system captures the physical environments in which HRI takes place, such as homes, museums, libraries, or any other relevant setting. This system sets the stage for HRI interactions and shapes the nature of robot use within specific environments.
Examples from our review include: robots in the home \cite{chen2022designing, armstrong2021longing, ahtinen2023robocamp, pelikan2019stubborn, levinson2022living} or surveys for in-home use~\cite{lin2021parental, lee2022can}; museums~\cite{mchugh2021unusual}; hospital~\cite{amirova2023effects}; or robots used in multiple contexts, between a playroom and non-co-located family members~\cite{abe2018chicaro}, a hospital and home~\cite{henry2020qualitative}, or classroom and home~\cite{de2021exploring}.

\subsection{Influences on Family-Robot Systems}
All three of these systems are bounded by four dimensions: roles assumed within a system; goals set within a system; processes followed within and between systems; and the changes in time.
 
\subsubsection{Roles}
The \textit{roles of family members} in child-robot interactions span across varying levels of involvement such as: \textit{direct} participants (\cite{chen2022designing, abe2018chicaro, armstrong2021longing, ahtinen2023robocamp, pelikan2019stubborn, mchugh2021unusual, henry2020qualitative, amirova2023effects, levinson2022living}), for example in parent-child co-learning and co-diary keeping~\cite{ahtinen2023robocamp}, parent-child storytelling~\cite{chen2022designing}, or parent-child goal-oriented play~\cite{mchugh2021unusual}; or \textit{indirect} influences on children's perceptions towards robots (e.g., ~\cite{de2021exploring}), including roles such as an informer, feedback provider, interpreter, consent provider, survey filler (e.g., ~\cite{lin2021parental, lee2022can}), diary keeper, reporter, passive observer.
Robots can assume various roles within the family, such as a companion for the child or an assistant for the parents~\cite{cagiltay2020investigating}. In the reviewed work, \textit{robot roles} span across storytelling companions~\cite{chen2022designing, lin2021parental}, playmates~\cite{abe2018chicaro, pelikan2019stubborn, mchugh2021unusual, levinson2022living}, learning companions~\cite{armstrong2021longing, ahtinen2023robocamp, de2021exploring, amirova2023effects}, childcare assistant~\cite{lee2022can}, and telepresence agent~\cite{henry2020qualitative}.

\subsubsection{Goals}
The \textit{goals set by individual family members}, for example, parents' desires for childcare support~\cite{lee2022can} or children's need for companionship~\cite{de2021exploring}, may influence the broader family acceptance and engagement with robots. As robots become more prevalent among family homes, conflicts and tensions regarding the use of these technologies will likely emerge due to the family members' multi-faceted needs (e.g., as seen in family interactions with smart home devices~\cite{beneteau2019communication, beneteau2020parenting, sun2021child}). 
The \textit{goals of robots} are inherently shaped by its use cases (e.g., storytelling, play, education) and may include goals such as sustaining long-term engagement, facilitating child-parent co-play, or a medium to connect distant family members.

\subsubsection{Processes}
The processes represent the actions that occur between systems or within a system. Take the example of a storytelling robot (robot system) placed in a library (contextual system) interacting with children and grandparents (family system). The \textit{processes between systems} may include the direct interactions between the child-grandparent-robot and library staff (e.g., turn-taking), as well as the indirect interactions between the library context and robot (e.g., the library's noise level impacting the robot's ability for speech recognition). The \textit{processes within a system} may capture the direct actions between the child and grandparent, or indirect actions such as the robot's connection to a cloud server. Inherently, the processes capture beyond a single dimension, and are shaped by the roles, goals, and affordances of all three systems.

\subsubsection{Time} 
The factor of time is crucial for a holistic understanding of family-centered HRI. It accounts for changes that happen over time and reflects the importance of long-term interactions.
Our review included several examples that captured families' long-term interactions with robots, which typically ranged between one week to one month~\cite{amirova2023effects, henry2020qualitative, levinson2022living, chen2022designing, ahtinen2023robocamp, pelikan2019stubborn} and in one case up to three months~\cite{armstrong2021longing}. 
The time dimension cuts across all three systems.
The indirect impact of time on the family system becomes visible in long-term interactions with robots, for example, uncovering improvements in family connections, changes in child development and learning, and shifts in family dynamics. 
Similarly, this dimension can capture the evolution of the robot's role and capabilities within the family over extended periods. For example, a robot's role may initially be an assistant for the family and, over time, may evolve into a companion for an older adult in the household. On the contrary, families may become less attached to the robot over time and have decreased engagement due to changes in the family system. In response, this may call for proposing novel and adaptive robot behaviors.
Nonetheless, contextual changes that happen over time may inevitably impact the robot and family system. For example, it may be easier for families to maintain routines with robots during the school year compared to the summer vacation.

\section{Discussion}
In this work, we introduced a theoretically motivated framework as a step toward family-robot interactions. We propose that, a family-robot interaction involves three systems: family system, robot system, and contextual system. Within this framework, it is necessary to explore, identify, and define the roles, goals, and processes between members involved in each system to understand the holistic nature of family-robot interactions. This framework was constructed given insights drawn from reviewed literature in child-robot interaction and insights from family theories. In our narrative review, we explored the current space of child-robot interactions that expand to broader family involvement. We found three gaps in the field: first, the limited body of work that focuses on contexts beyond therapy and healthcare; second, the lack of representation of family members involved in child-robot interaction; and third, limited exploration of long-term social robot use that can adapt to changes in roles and dynamics within the family.

With further synthesis, we identified two core themes that capture the high-level interplay between family members and robots. First, robots have the potential to support the needs and preferences of children and their surrounding care ecosystems, embodying roles that could enhance family connections. They can serve as catalysts for family togetherness, create shared experiences, and promote lasting memories through collaborative activities.
Second, family characteristics, such as parents' STEM backgrounds, employment, individual beliefs, values, and communication patterns, play a pivotal role in shaping how family members perceive and interact with robots, potentially influencing long-term acceptance and adoption. By understanding the diversity among family characteristics and the unique needs of individuals within the family, researchers can inform design decisions for robot capabilities that are more suitable to real-world use-cases.
Finally, we translated these insights to construct a family-centered framework for family-robot interactions. 

\subsection{Practical Applications of the Framework} 
Here, we discuss strategies for validating and applying the family-centered framework for family-robot interactions and ways in which the framework may provide value to its practitioners. First, to ensure the holistic understanding of family-robot interactions, it becomes imperative to unravel and define the roles, goals, and processes within each system.
\textit{To validate the framework}, we propose engaging in empirical studies to assess the extent to which the framework explains observed family-robot interactions. For example, a study may explore how family members’ characteristics and perceived trust may influence turn-taking in family-robot group interactions and inform robot design. For validation, one should demonstrate the emergence of systems and boundaries as outlined in the framework, providing empirical support and enhancing our comprehension of family-robot interactions. 
Once validated, the framework aims to propose value as a versatile tool for HRI researchers seeking to take a family-centered lens. This would enable \textit{applying the framework,} which would involve guiding researchers in the early stages of their work, aiding them in making informed technical design decisions, structuring study designs, formulating hypotheses, selecting study sites, and recruiting participants. By integrating this family-centered perspective into the research process, the framework should ideally ensure that the study captures the nuanced interplay between family members and robots. For example, the framework might serve as a template to create a family-robot integration plan, specifying design requirements tailored to family needs and robot affordances.
Moreover, there is potential for \textit{customization and extension} of the framework based on specific research contexts and new insights. Researchers are encouraged to adapt the framework to suit the unique characteristics of their study populations and contexts. As insights emerge from varying family structures, cultural contexts, and long-term implications, the framework will likely serve as a basis to be customized and expanded to align with the evolving landscape of family-robot interactions.

\subsection{Limitations and Future Directions}
We acknowledge several limitations in our work. First, our review process for constructing the framework does not constitute as a comprehensive examination of existing literature. Thus, our synthesis may exclude some studies, viewpoints, or nuanced findings. The selected papers for synthesis were included as a first step toward a framework demonstrating how child-robot interaction can involve other family members. Future research must prioritize design, user evaluations, and extensive scoping reviews of relevant family-robot interaction phenomena to better understand the landscape and expand our framework.
Furthermore, it is worth noting that many of the discussion points included in this framework are already known challenges within HRI research. Long-term interaction, adaptation, personalization, cultural diversity, and ethical and social implications for children have long been recognized as critical areas of exploration. Our framework situates these dimensions in the broader family-robot interaction context. 
We believe that focusing on families as a whole may provide a more holistic approach to addressing broader challenges in HRI research. The framework aims to increase HRI researchers' participation in these important conversations around robots tailored for families. Despite these limitations, we aim for our work to serve as a valuable starting point for future research directions toward family-centered HRI, fostering innovation and collaboration within the field.

\section{Conclusion}
We presented a theoretically motivated framework informed by family theories and the growing literature in HRI focusing on children and families. We argue that a family-centered approach in HRI should encapsulate the multifaceted interactions between family members and robots. We aim that our proposed framework may serve as a first step toward designing for family-robot interactions.

\begin{acks}
This work was supported by NSF awards \#2312354 and \#2247381. We would like to thank Pragathi Praveena, Hailey Johnson, Casey Wong, and the anonymous reviewers for their constructive and actionable feedback that helped improve the quality of our paper.
\end{acks}

\bibliographystyle{ACM-Reference-Format}
\balance
\bibliography{references}

\end{document}